\def\year{2022}\relax
\documentclass[letterpaper]{article} 
\usepackage{aaai22}  
\usepackage{times}  
\usepackage{helvet}  
\usepackage{courier}  
\usepackage[hyphens]{url}  
\usepackage{graphicx} 
\usepackage{natbib}  
\usepackage{caption} 
\DeclareCaptionStyle{ruled}{labelfont=normalfont,labelsep=colon,strut=off} 
\usepackage{algorithm}

\usepackage{newfloat}
\usepackage{listings}
\floatstyle{ruled}
\newfloat{listing}{tb}{lst}{}
\floatname{listing}{Listing}

\usepackage[utf8]{inputenc} 
\usepackage[T1]{fontenc}    
\usepackage{hyperref}       
\usepackage{url}            
\usepackage{booktabs}       
\usepackage{amsfonts}       
\usepackage{nicefrac}       
\usepackage{microtype}      
\usepackage{xcolor}         

\usepackage{amsmath,amsopn,amssymb}
\usepackage{graphicx,xspace,color,soul}
\usepackage{algorithm,algorithmicx,algpseudocode}
\usepackage{epsfig}
\usepackage{longtable,multirow}
\usepackage{array}
\usepackage{bm,epstopdf,booktabs}
\usepackage{subfig}
\usepackage{enumitem}
\usepackage{bbm}
\usepackage{wrapfig}

\def \ie{{\it i.e.}}
\def \eg{{\it e.g.}}

\newlength\savewidth\newcommand\shline{\noalign{\global\savewidth\arrayrulewidth
  \global\arrayrulewidth 1pt}\hline\noalign{\global\arrayrulewidth\savewidth}}

\makeatletter\renewcommand\paragraph{\@startsection{paragraph}{4}{\z@}
  {.5em \@plus1ex \@minus.2ex}{-.5em}{\normalfont\normalsize\bfseries}}\makeatother

\title{Not All Voxels Are Equal: Semantic Scene Completion \\ from the Point-Voxel Perspective}

\author{
Xiaokang Chen\textsuperscript{\rm 1\thanks{Equal contribution. X. Chen and J. Wang designed the method. J. Tang and X. Chen performed the algorithm verification and co-wrote the manuscript. All the authors discussed the results and commented on the manuscript.}}, 
Jiaxiang Tang\textsuperscript{\rm 1\footnotemark[1]},
Jingbo Wang\textsuperscript{\rm 2\footnotemark[1]},
Gang Zeng \textsuperscript{\rm 1}
}
\affiliations{
\textsuperscript{\rm 1}School of Intelligence Science and Technology, Peking University,~ \textsuperscript{\rm 2}Chinese University of Hong Kong\\
\{pkucxk,tjx,zeng\}@pku.edu.cn,~ wj020@ie.cuhk.edu.hk\\
}

\usepackage{bibentry}

\begin{document}

\maketitle


\begin{abstract}
We revisit Semantic Scene Completion (SSC), a useful task to predict the semantic and occupancy representation of 3D scenes, in this paper. A number of methods for this task are always based on voxelized scene representations for keeping local scene structure. However, due to the existence of visible empty voxels, these methods always suffer from heavy computation redundancy when the network goes deeper, and thus limit the completion quality.
To address this dilemma, we propose our novel point-voxel aggregation network for this task. Firstly, we transfer the voxelized scenes to point clouds by removing these visible empty voxels and adopt a deep point stream to capture semantic information from the scene efficiently. Meanwhile, a light-weight voxel stream containing only two 3D convolution layers preserves local structures of the voxelized scenes. Furthermore, we design an anisotropic voxel aggregation operator to fuse the structure details from the voxel stream into the point stream, and a semantic-aware propagation module to enhance the up-sampling process in the point stream by semantic labels. We demonstrate that our model surpasses state-of-the-arts on two benchmarks by a large margin, with only depth images as the input.
\end{abstract}

\section{Introduction}

\begin{figure}[htbp]
  \centering
  \includegraphics[width=\linewidth]{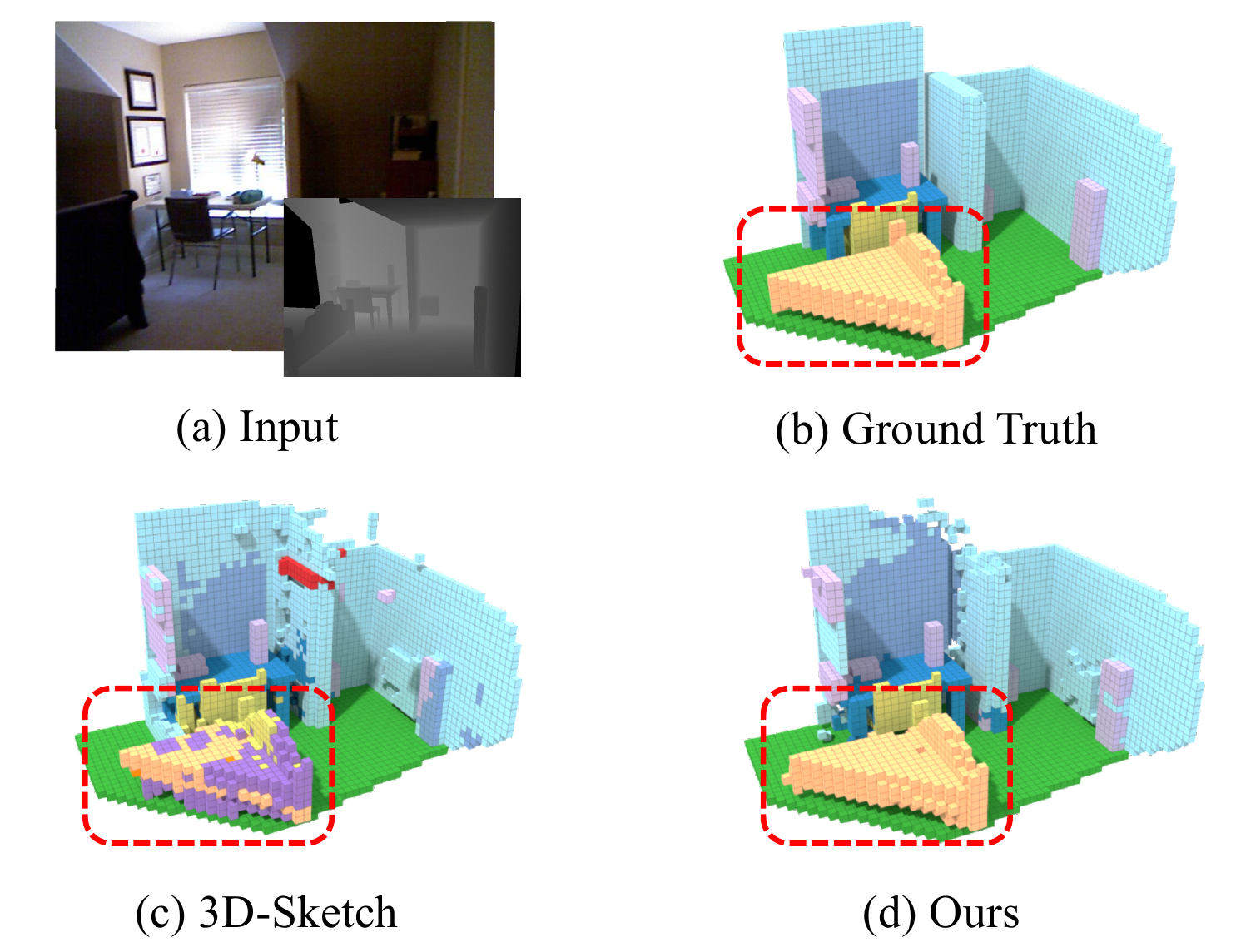}
  \caption{\textbf{Visualization of Semantic Scene Completion task}. 
  Our method generates more reasonable results compared to 3D-Sketch~\cite{chen20203d} while significantly lowering the computational costs.}
  \label{fig:teaser}
\end{figure}

With a partial 2D observation, humans are capable of understanding the 3D space and inferring the objects behind the occlusion. 
Similarly, the capability to capture the structure and semantic information of 3D scenes is beneficial for many real-world applications, including robotics, virtual reality, and interior design. 
To achieve this goal, we need to perform scene completion and scene labeling tasks, which are proved to be closely correlated by~\cite{song2017semantic-sscnet}.
Semantic Scene Completion (SSC) is therefore put forward to predict 3D geometry and semantics simultaneously from a partial observation, which is an emerging topic in recent years.

Previously, most methods~\cite{song2017semantic-sscnet,guo2018view-vvnet,garbade2018two-ts3d,liu2018see-satnet} solve this challenging problem with voxelized partial observations.
To handle these voxelized scenes, 3D convolution networks are always adopted by these methods to learn the occupancy and semantic information of each voxel.
Although voxel representations preserve abundant structure details of the partial 2D observation, \textit{not all voxels are of equal importance in this volume}.
In particular, there exist lots of visible empty voxels (\eg, atmosphere in the visible region) in the voxelized SSC data by nature.
These voxel-based methods have to perform unnecessary calculations on them in forward propagation, but ignore them in backward propagation since the labels are already known.
Therefore, they always suffer from heavy computation redundancy, especially for trying to keep the high scene resolution in a deep 3D convolution network.
To solve this problem, an efficient sparse data structure should be used, such as the point cloud or voxel octree~\cite{liu2020neural,takikawa2021neural}.
An early attempt~\cite{zhong2020spcnet} removes these visible empty voxels and adopts a point cloud based network to extract features from this non-grid data for SSC task,
but it is intrinsically weak in local structure modeling since the point cloud representation is sparse.

Therefore, it is crucial to consider the complementarity between voxel-based and point-based scene representation into the SSC framework. 
Unlike most other 3D computer vision tasks, such as 3D detection or segmentation, the data for SSC tasks is usually voxelized since the goal is to predict the semantic and occupancy of voxels in this 3D scene. 
The point clouds are extracted from these voxels. 
In fact, the voxel representation is denser than the point representation in our setting, for that only a part of point clouds are sampled as input during training. 
Thus, we design our Point-Voxel Aggregation Network (PVA-Net), where two 3D convolution layers keep the details from the voxelized scenes and a deep point cloud based network captures semantic information efficiently.

We adopt the point stream as the main stream of our network for its low memory requirement. 
Meanwhile, a light-weight voxel stream is used to extract structure details, which acts as a complement to the point stream.
To efficiently fuse the point stream and the voxel stream, we propose a novel Anisotropic Voxel Aggregation (AVA) module to aggregate information in voxels for each center point. 
Given the position of a center point, we apply three ellipsoidal receptive fields to extract feature patterns from the voxels in different directions and concatenate them with the center point's features.
Moreover, we design a semantic-guided decoder that consists of several Semantic-aware Propagation (SP) modules, which encourages feature propagation between points belonging to the same semantic class.

We summarize our contributions as follows:
\begin{itemize}
    \item To avoid the redundant computation in visible empty voxels in the SSC task, we convert the valid volume data to points and introduce the Point-Voxel Aggregation Network, which combines the low memory requirement of point-based methods and the local structure modelling ability of voxel-based methods.
    \item We propose the Anisotropic Voxel Aggregation module to efficiently fuse the structure information from a light-weight voxel stream into the point stream, and the Semantic-aware Propagation module to encourage feature propagation between points of the same semantic class.
    \item Our method outperforms state-of-the-arts by a large margin on two public benchmarks, with only depth images as the input.
\end{itemize}

\section{Related Work}

\subsection{Deep Learning for 3D Scene Analysis}
Learning semantic information of given scene with 3D information has drawn increasing attention in recent years.
Rather than directly using 3D data, previous methods always focus on RGBD images~\cite{xing20192, xing2019coupling, xing2020malleable, chen2020bi} to understand the semantic information of the given scene.
Different from 2D images, 3D data have various data representations such as voxels and point clouds and can 
facilitate various different applications~\cite{wang2021scene, rong2021frankmocap, rong2021monocular,JiaxiangTang2022CompressiblecomposableNV,JiaxiangTang2023PointSU,JiaxiangTang2023DelicateTM}. 
Lots of methods have been proposed to handle different representations.
3D CNNs are the straightforward extension of 2D CNNs to 3D voxels. 
Early researches~\cite{chang2015shapenet,Wu20153DSA,maturana2015voxnet,Zhou2018VoxelNetEL} rely on 3D convolutions to process 3D voxel data in regular grids.

Point-based methods learn 3D point cloud representations directly by defining permutation-invariant point convolutions in irregular space. 
PointNet~\cite{qi2017pointnet} first uses a shared MLP on every point individually followed by global max-pooling to extract global features. 
Pointnet++~\cite{qi2017pointnet++} introduces hierarchical architectures to learn local features and increases modal capacity. 
Later works~\cite{Rethage2018FullyConvolutionalPN,Landrieu2018LargeScalePC,Wu2018SqueezeSegCN,Zhao2019PointWebEL,Milioto2019RangeNetF,Komarichev2019ACNNAC,Lang2019PointPillarsFE,Hu2020RandLANetES,xu2021paconv} focus on more effective and general point operations, such as explicit point convolution kernels where the weights can be directly learned without intermediate MLP representations~\cite{hua2018pointwise,Li2018PointCNNCO,thomas2019kpconv,lin2020fpconv} and graph convolutions~\cite{Wu2019PointConvDC,Li2019DeepGCNsCG,Wang2019GraphAC}.

\begin{figure*}[ht]
    \centering
    \includegraphics[width=0.9\textwidth]{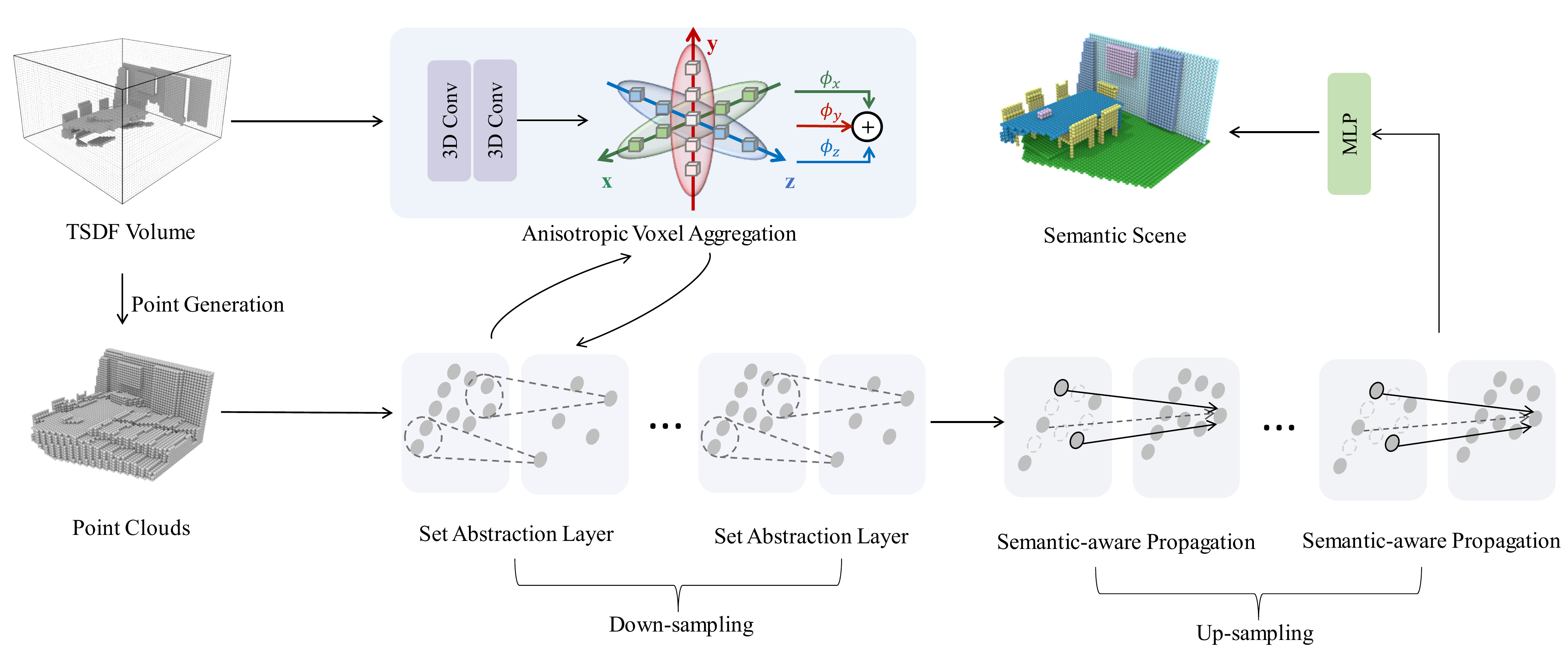}
    \caption{\textbf{The overall architecture of the proposed method}. 
    We generate point clouds from input TSDF volumes and use an encoder-decoder architecture to predict the semantic labels, 
    with an Anisotropic Voxel Aggregation module to aggregate the local structure information from the voxels.
    }
    \label{fig:network}
\end{figure*}
Multi-modality fusion is also a long-term problem in 3D deep learning. 
Recently, a few works begin to leverage the advantages of point cloud and voxel representation together in deep neural networks. 
PV-CNN~\cite{liu2019point} is proposed to represent 3D data in sparse points to save memory cost and perform convolutions in voxels to obtain the contiguous memory access pattern.
PV-RCNN~\cite{Shi2020PVRCNNPF} defines a Voxel Set Abstraction (VSA) module to summarize voxel features into key points to further explore this problem.
However, this work focuses on object detection and relies on a heavy voxel stream to regress object proposals.
To improve the learning efficiency and capability of the framework, we differ from the design choices in PV-RCNN as follows: 
1) We only use a light-weight voxel stream to assist the main point stream; 
2) Our AVA module is more general than the VSA module, which can be seen as a special case of our AVA module using spherical receptive fields; 
3) We focus on scene completion and propose the SP module to aggregate features at different stages with semantic guidance during up-sampling, which is not considered by PV-RCNN. 

\subsection{Semantic Scene Completion}
Semantic Scene Completion (SSC) aims to predict a complete voxel representation of a 3D scene and each voxel's semantic label, usually from a single-view depth map observation.
SSCNet~\cite{song2017semantic-sscnet} first combines semantic segmentation and scene completion in an end-to-end way, showing that the two tasks are highly coupled and can be learned together to improve the performance. 
ESSCNet~\cite{zhang2018efficient-esscnet} introduces Spatial Group Convolution (SGC) which divides the voxels into different groups to save computational cost. 
Later works~\cite{guo2018view-vvnet,wang2019forknet,zhang2019cascaded-ccpnet,chen2020real} improve the performance with better architecture, such as 2D-3D combination, cascaded context pyramid and so on.
Guedes \textit{et. al.}~\cite{Guedes2018SemanticSC} first investigates the potential of the RGB images to improve SSCNet. 
After that, many methods~\cite{garbade2018two-ts3d,liu2018see-satnet,dourado2019edgenet,li2019rgbd-ddrnet,li2020attention,chen20203d} take RGB images as an additional input with the depth map and explore the complementarity between the two modalities. Most recent work~\cite{li2021imenet} further explores the interaction between 2D segmentation and 3D SSC, but they rely on heavy networks to perform the feature extraction.

These methods all utilize 3D CNNs as the backbone, causing unnecessary computational cost in the visible empty voxels.
Whereas AIC-Net~\cite{li2020anisotropic} also proposes an anisotropic convolution to model the voxel-wisely dimensional anisotropy, our AVA module differs in both formulation and functionality, focusing on synchronizing features from voxels to point representation with a more flexible ellipsoidal receptive field.

Different from those voxel-based methods above,
SPCNet~\cite{zhong2020spcnet} first introduces a point-based network to address the SSC problem, by training a point network on the observed points and then obtaining the features of the occluded points through bilinear interpolation. 
Due to the interpolation process that depends on distance metrics, the occluded points far from the visible points are hard to predict in their method.
Furthermore, point-based method alone is not enough to retain detailed local structure information during the down-sampling progress.
Therefore, we draw ideas from the Point-Voxel methods and propose a two-stream network, where an efficient point stream extracts semantic features and a light-weight voxel stream provides dense local structure information through the AVA module. 

\section{Methodology}
\subsection{Overview of the Proposed Method}

The overall architecture of the proposed method is illustrated in Figure~\ref{fig:network}. 
Our model consists of a point stream and a voxel stream.
To reduce the computation redundancy, we convert the TSDF volume to a point cloud by removing the visible empty voxels, which serves as the input of the point stream. 
The point stream adopts a PointNet++~\cite{qi2017pointnet++}-like encoder-decoder architecture.
The encoder extracts the semantic features in a hierarchical way and the decoder encourages feature propagation in points of the same class. 
Meanwhile, a light-weight voxel stream that only contains two 3D dense convolutional layers is applied to the TSDF volume.
An AVA module is proposed to aggregate the local voxel features to the point features. 
Finally, the predicted point labels are converted back to the voxel representation to calculate the evaluation metrics. 

\subsection{Point Clouds Generation}
\label{sec:point-encoding}
\begin{figure}[htbp]
  \centering
  \includegraphics[width=0.6\linewidth]{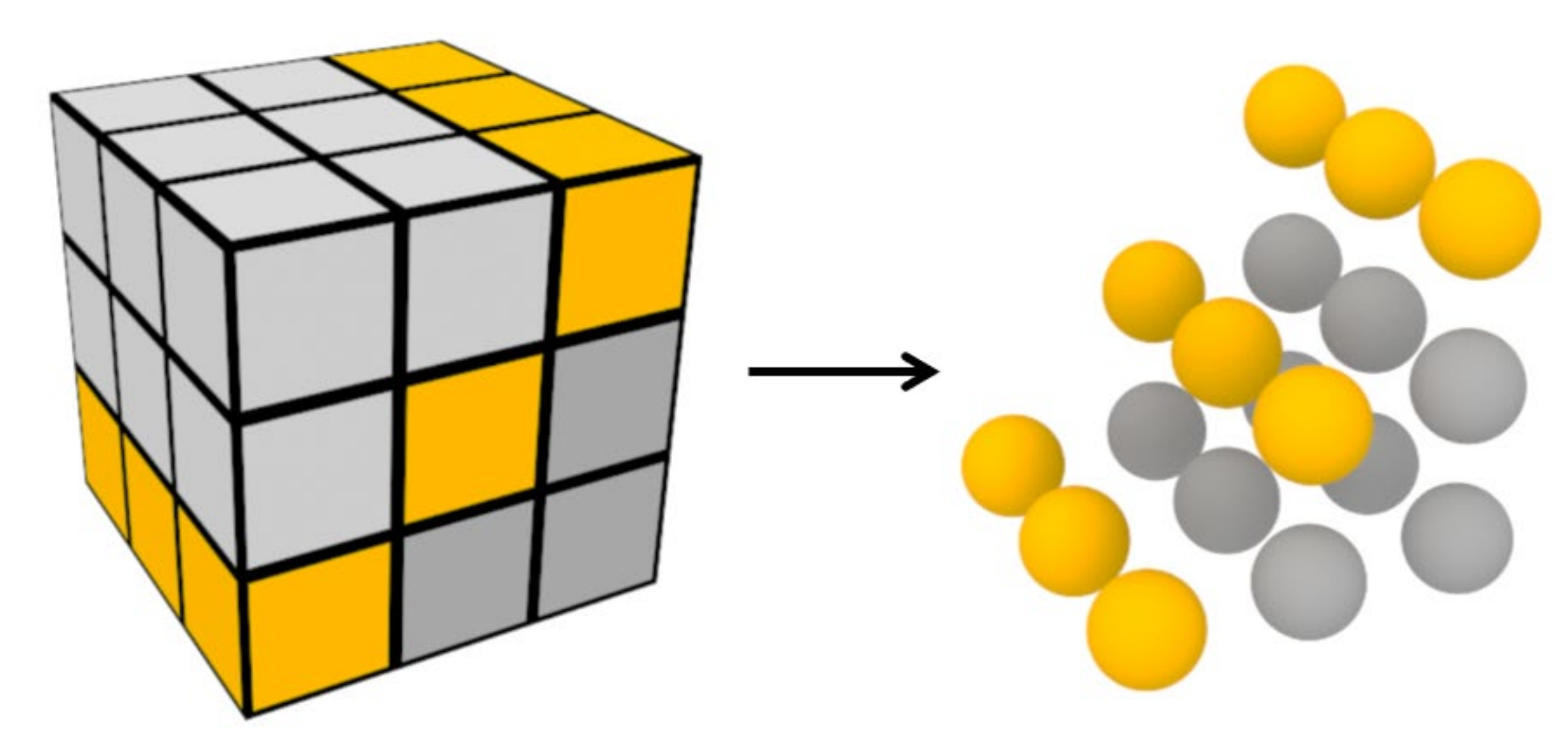}
  \caption{
 \textbf{Points generation from the voxel volumes}. 
 Only the observed surface (yellow) and occluded regions (dark gray) are kept, while the visible empty voxels (light gray) are discarded.
 For example, the average number of the kept points is 16313 for the 60 $\times$ 36 $\times$ 60 input voxels from the NYUCAD dataset, which means about 87\% of the input voxels are redundant.
 }
  \label{fig:Point_gen}
\end{figure}

Voxel-based methods always encode the depth map into a 3D TSDF volume~\cite{song2017semantic-sscnet}, and carry out later procedures in this voxel space. 
However, we argue that not all voxels in this volume are of equal importance for the SSC task.
In fact, there are three kinds of voxels inside this volume:
1) the \textbf{observed surface} voxels which are directly projected from the given depth image,
2) the \textbf{occluded} voxels behind the observed surface which we need to complete and recognize,
and 3) the \textbf{visible empty} voxels (such as the atmosphere) between camera and the observed surface. 
The last kind of voxel is useless for our task since we already know it's empty. 
Thus, these voxels are removed during our point clouds generation process.
As shown in Figure~\ref{fig:Point_gen}, we only keep the observed surface and the occluded regions in our point clouds.

Each input point $\mathbf{p}_i$ has a $5$-dim feature vector $\mathbf{f}_i = (x_i, y_i, z_i, t_i, h_i)$. 
Suppose $\mathbf{p}_i$ is generated from voxel $\mathbf{v}_i$ in the $60 \times 36 \times 60$ 3D volume, then $x_i, y_i, z_i$ are normalized $x$-$y$-$z$ indexes of $\mathbf{v}_i$ in the volume. 
$t_i$ is the TSDF value of $\mathbf{v}_i$, and $h_i$ is the normalized height value of $\mathbf{v}_i$. 
Please note that $x_i, y_i, z_i$ are normalized according to the mass center of the points in the scene, while $h_i$ is normalized by $36$, the maximum height of the voxelized scene. 
We think the normalized height serves as a prior that describes the positions of objects in the room. 
This could help to distinguish some categories with significantly different height values, such as the floor and the ceiling.

\subsection{Anisotropic Voxel Aggregation}
Due to the sparsity of the point clouds, it is hard for the point stream to model the detailed structure information which is important for the scene completion and recognition. 
Since we have the denser volume data as well, we design a voxel stream to extract the structure features and propose the Anisotropic Voxel Aggregation (AVA) module to fuse the point-voxel features.

We first extract the local features of the TSDF volume through two simple convolution layers. 
This requires little computational cost and takes about only 15.0\% of the overall memory cost, but enables each voxel in the volume to have a suitable receptive field to encode local geometry information. 
As shown in the top part of Figure~\ref{fig:network}, for each center point $\mathbf{p}_i = (x_i, y_i, z_i)$ in the point cloud, we define three ellipsoidal receptive fields with $x, y, z$ axis as the major axis respectively. 
Taking the $x$-axis as an example, the receptive field $\mathcal N_x(i)$ of $\mathbf{p}_i$ in the volume could be defined as:
\begin{equation}
\resizebox{.85\linewidth}{!}{$
    \displaystyle
    \mathcal N_x(i) = \left\{ \mathbf{v}_j \middle| \frac {(x_j - x_i)^2} {(kr)^2} + \frac {(y_j - y_i)^2} {r^2} + \frac {(z_j - z_i)^2} {r^2} < 1 \right\}
$}
\end{equation}
where $\mathbf{v}_j$ is the $j$-th voxel and $(x_j, y_j, z_j)$ is its position, $r$ is the radius for minor axes, and $k > 1$ is a scale factor for the major axis. 
Unless mentioned specifically, we use $3$ as the default value for $k$. 
The receptive field along the $y$-axis and $z$-axis could be defined in a similar way. 
From the perspective of pattern recognition, the anisotropic receptive field ensures us to activate feature patterns in three directions, which is more flexible and effective than the isotropic spherical receptive field.

\begin{figure}[htbp]
  \centering
  \includegraphics[width=\linewidth]{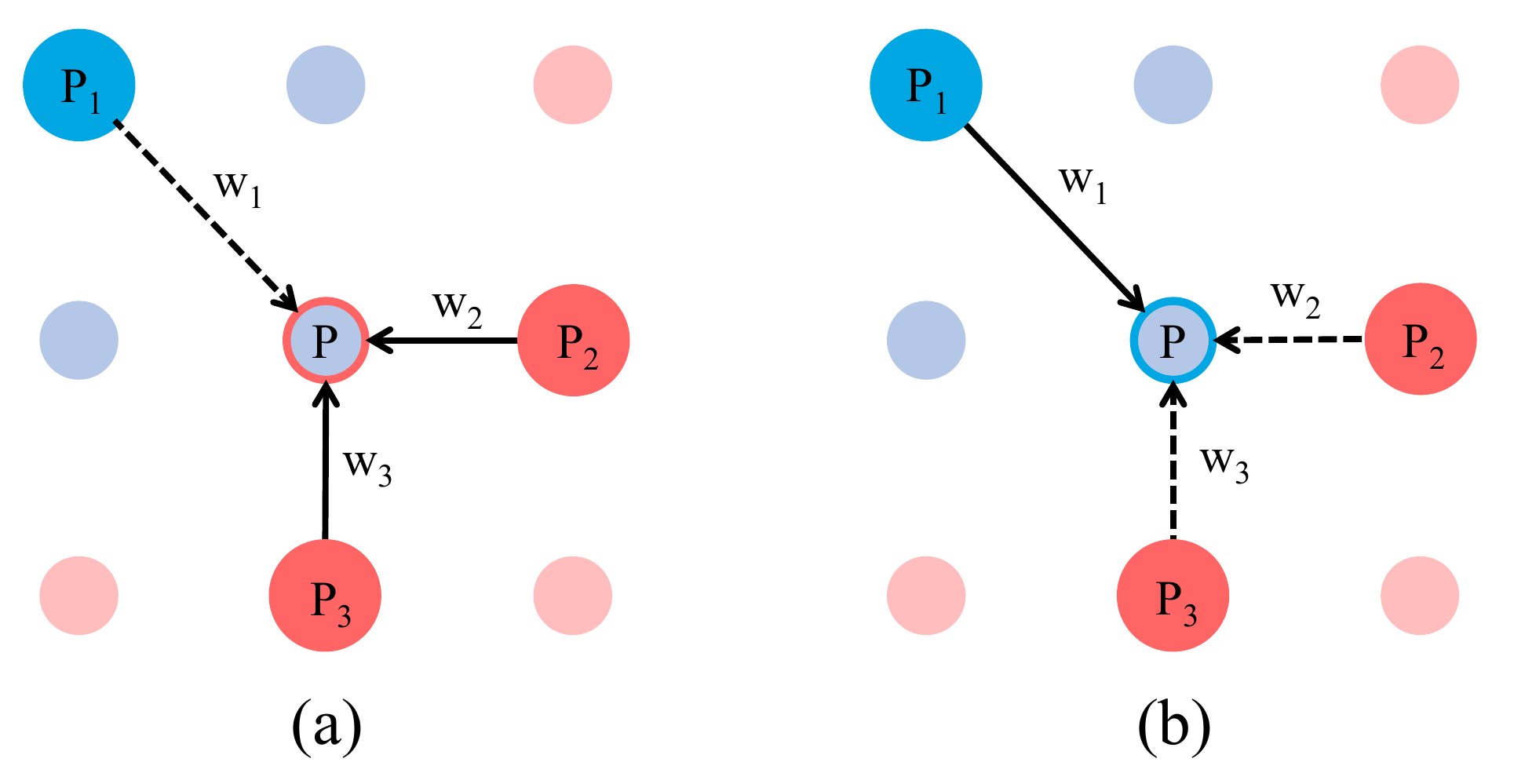}
  \caption{ \textbf{Different Feature Propagation strategies.} (a)  Feature Propagation (FP) in PointNet++~\protect\cite{qi2017pointnet++}. (b) The proposed Semantic-aware Propagation (SP). Different colors mean different semantic classes. Larger points are from a deeper layer, while smaller points are in the current layer. Dashed lines mean smaller weight. The boundary color of the center point means the interpolated features are dominated by which semantic class. In (a), the center point is dominated by the wrong class due to the unsuitable FP strategy, while the proposed method (b) avoids this problem.}
  \label{fig:SP}
\end{figure}

Then we could aggregate the structure features of voxels around $\mathbf{p}_i$ with: 
\begin{align}
    \mathbf{f}^{\text{fuse}}_{i}
    = \sum_{d\in \{x, y, z\}}  \max_{j \in \mathcal N_d(i)} \{\phi_{d}(\mathbf{f}^{\text{point}}_i, \mathbf{f}^{\text{voxel}}_j)\}
\end{align}
where $\mathbf{f}^{\text{fuse}}_{i}$ is the fused point-voxel feature, 
$\mathbf{f}^{\text{point}}_i$ is the feature of $\mathbf{p}_i$, 
$\mathbf{f}^{\text{voxel}}_j$ is the feature of $j$-th voxel in the volume,
$\phi_{d}$ is a MLP layer for non-linear feature extraction in the $d$-axis,
$\max$ denotes max-pooling operation that keeps the maximum activation in the neighborhood,
and $\mathcal N_d(i)$ represents the set of neighbor voxels of $\mathbf{p}_i$ inside the ellipsoidal receptive field with the $d$-axis as the major axis.
This AVA module enables the sparse center points to aggregate local structure information from nearby dense voxels.
Therefore, the information from voxels could positively affect the completion and recognition of point clouds through back propagation.

\subsection{Semantic-aware Propagation}
\label{sec:SP}
During the encoding process, Set Abstract (SA)~\cite{qi2017pointnet++} layers will down-sample the input points. 
Suppose we have $N$ SA layers in the network, named $\text{SA}^{(1)}$, ..., $\text{SA}^{(N)}$. 
We denote $\mathbf X^{(0)} = \{(\mathbf{p}^{(0)}_i, \mathbf{f}^{(0)}_i)\}$ as the raw input point set, where $\mathbf{f}^{(0)}_i$ is the feature of the point $\mathbf{p}^{(0)}_i$. 
Then the corresponding output point sets of $N$ SA layers are $\mathbf X^{(1)}$, ..., $\mathbf X^{(N)}$, respectively. 
Please note that if a point belongs to $\mathbf X^{(i)}$ ($i \ge 1$), then it must belong to $\mathbf X^{(i-1)}$ as well, because each SA layer only down-samples points from the former layer. 

To obtain the features of all the raw input points, 
we propose the Semantic-aware Propagation (SP) module, which is a hierarchical propagation strategy as shown in Figure~\ref{fig:SP}.
For a target point $\mathbf{p}^{(l)}_i$ in $\mathbf X^{(l)}$, we find its $k$-neighbors ($\mathbf{p}^{(l+1)}_1$, ..., $\mathbf{p}^{(l+1)}_k$) in $\mathbf X^{(l+1)}$ according to the $xyz$ coordinates. 
To interpolate the feature of $\mathbf{p}^{(l)}$, the general feature propagation can be represented by:
\begin{equation}
\label{eq:edge-weight}
\mathbf{f}^{(l)}_i = \frac {\sum_{j \in \mathcal N_k^{(l+1)}(i)}w_{i,j}^{(l)} \mathbf{f}_j^{(l+1)}} {\sum_{j \in \mathcal N_k^{(l+1)}(i)}w_{i,j}^{(l)}}
\end{equation}
where $\mathcal N_k^{(l+1)}(i)$ is the set of $k$ nearest neighbors of $\mathbf{p}^{(l)}_i$ in $\mathbf X^{(l+1)}$, and $w_{i,j}^{(l)}$ is the weight factor for $\mathbf{f}^{(l+1)}_j$ with respect to the point $\mathbf{p}^{(l)}_i$.

An intuitive idea is to measure the similarity between the point $\mathbf{p}_i^{(l)}$ and $\mathbf{p}^{(l+1)}_j$ for $j \in \mathcal N_k^{(l+1)}(i)$ and use the similarity as the weight factor. 
However, since $\mathbf{p}_i^{(l)}$ and $\mathbf{p}^{(l+1)}_j$ belong to different levels and thus are embedded to different feature spaces, 
it is not suitable to directly compare their feature vectors. 
We notice that the point $\mathbf{p}^{(l+1)}_j$ also exists in $\mathbf X^{(l)}$ since the points in $\mathbf X^{(l+1)}$ is a subset of $\mathbf X^{(l)}$. 
Then we could measure the similarity between $\mathbf{p}_i^{(l)}$ and $\mathbf{p}^{(l)}_j$ in a learnable manner:
\begin{equation}
\label{eq:pair-sim}
    w_{i,j}^{(l)} = \sigma(\phi(\mathbf{f}_i^{(l)} || \mathbf{f}_j^{(l)}))
\end{equation}
where $\sigma$ is the sigmoid function, $\phi$ is a MLP and $||$ means channel-wise concatenation. 

In this way, we could interpolate point features with semantic information, which is helpful in SSC task.
We explicitly supervise the learned weights during training, 
by setting the ground truth of $w_{i,j}^{(l)}$ to $1$ if the two points belong to the same semantic class, and $0$ if they belong to different classes.
We think this could encourage the network to only propagate semantically similar features, which weakens the effect of neighbor points from different classes during interpolation.

\subsection{Training Loss}

The training loss involves two terms: SSC loss $\mathcal L_{\text{SSC}}$ and SP loss $\mathcal L_{\text{SP}}$.
The SSC loss is a weighted voxel-wise cross-entropy loss:
\begin{equation}
    \displaystyle
    \mathcal L_{\text{SSC}} = \frac {1}{N_{\text{valid}}} \sum_{i,j,k} m_{i,j,k} \mathcal L_{\text{CE}} (p_{i,j,k}, \,y_{i,j,k})
\end{equation}
where $m_{i,j,k}$ is set to $1$ if the voxel at index $(i,j,k)$ is not visible empty (\ie, can be converted to a point) or $0$ otherwise. 
$y_{i,j,k}$ is the ground truth label,
$p_{i,j,k}$ is the prediction of the voxel mapped back from the corresponding point,
$N_{\text{valid}} = \sum_{i,j,k} m_{i,j,k}$ is the number of valid voxels in this volume, 
and $\mathcal L_{\text{CE}}$ is the cross-entropy loss.

The SP loss is designed to supervise the pairwise similarity introduced in Equation \ref{eq:pair-sim}. The SP loss could be formulated as:
\begin{equation}
    \mathcal L_{\text{SP}} = \frac {1}{N_{\text{pairs}}} \sum_{l} \sum_{0 \leq i \leq \lvert\mathbf{X}^{(l)}\rvert} \sum_{j \in \mathcal N_k^{(l+1)}(i)} \mathcal L_{\text{CE}} (w_{i,j}^{(l)}, \,\mathbf{G}_{i,j}^{(l)})
\end{equation}

where $N_{\text{pairs}}$ represents the number of point pairs involved, 
$\smash\lvert\mathbf{X}^{(l)}\rvert$ means the number of points in the $l$-th level, 
${\mathcal N_k^{(l+1)}(i)}$ is defined in Section \ref{sec:SP}. 
${\mathbf{G}_{i,j}^{(l)}}$ is the ground truth of the pair-wise similarity. If two points $\mathbf{p}_i^{(l)}$ and $\mathbf{p}_j^{(l)}$ belong to the same category, it is 1, otherwise it is 0.

We optimize the entire network by the balanced combination of the two terms:

\begin{equation}
\label{loss-func}
    \mathcal L = \mathcal L_{\text{SSC}} + \lambda \mathcal L_{\text{SP}}
\end{equation}

\section{Experiments}

In this section, we evaluate the proposed method and compare it with state-of-the-art methods on two public datasets, NYU~\cite{nyudv2} and NYUCAD~\cite{firman2016structured}.


\subsection{Datasets and Evaluation Metrics}
\label{sec:datasets}
\paragraph{Datasets.} 
The \textbf{NYU} dataset~\cite{nyudv2} consists of $1,449$ realistic indoor RGB-D scenes captured via a Kinect sensor~\cite{song2017semantic-sscnet}. 
Since the real-world completed scenes are hard to be captured, human annotations provided by~\cite{guo2015predicting} are widely used as the ground truth completion.
However, as discussed in \cite{song2017semantic-sscnet}, there exists many misalignments between the depth images and the corresponding 3D labels in the NYU dataset, 
which makes it hard to evaluate accurately. 
To solve this problem, the high-quality synthetic \textbf{NYUCAD} dataset is proposed by \cite{firman2016structured},
where the depth maps are projected from the ground truth annotations and thus avoid the misalignments. 
Following previous works~\cite{song2017semantic-sscnet,chen20203d,garbade2018two-ts3d}, we choose NYU and NYUCAD to evaluate our method.

\paragraph{Evaluation Metrics.} 
We follow \cite{song2017semantic-sscnet} to use precision, recall and voxel-level intersection over union (IoU) as the evaluation metrics. 
Two tasks are considered, namely, semantic scene completion and scene completion. 
For the task of semantic scene completion, we evaluate on both the observed surface and occluded regions and report the mIoU of each semantic class.
For the task of scene completion, we treat all non-empty voxels as class `1' and all empty voxels as class `0', and then evaluate the binary IoU on the occluded regions.

\begin{table*}[ht]
\begin{center}
\resizebox{\linewidth}{!}{
\begin{tabular}{l|c|c|c c c|c c c c c c c c c c c | c} 
\shline
& & & \multicolumn{3}{c|}{scene completion} & \multicolumn{12}{c}{semantic scene completion} \\ \hline
Methods  & Input & Resolution & prec. & recall & IoU & ceil. & floor & wall & win. & chair & bed & sofa & table & tvs & furn. & objs. & mIoU \\ 
\hline
Zheng $et\, al.$~\cite{zheng2013beyond} & D & (240, 60)	& 60.1 & 46.7 & 34.6 & - & - & - & - & - & - & - & - & - & - & - & - \\ 
Firman $et\, al.$~\cite{firman2016structured} & D & (240, 60)	& 66.5 & 69.7 & 50.8 & - & - & - & - & - & - & - & - & - & - & - & - \\ 
SSCNet~\cite{song2017semantic-sscnet} & D & (240, 60) & 75.4 & \textbf{96.3} & 73.2 & 32.5 & 92.6 & 40.2 &  8.9 & 33.9 & 57.0 & 59.5 & 28.3 &  8.1 & 44.8 & 25.1 & 40.0\\ 
CCPNet~\cite{zhang2019cascaded-ccpnet} & D & (240, 240) & {91.3} & 92.6 & 82.4 & 56.2 & \textbf{94.6} & 58.7 & \textbf{35.1} & 44.8 & 68.6 & 65.3 & 37.6 & 35.5 & 53.1 & 35.2 & 53.2 \\
SPCNet~\cite{zhong2020spcnet} & D & (60, 60) &  81.4 & 70.9 & 61.0 & 58.1 & 91.6 & 53.7 & 13.0 & 52.1 & 68.9 & 57.7 & 31.9 & 6.4 & 50.5 & 28.1 & 46.6 \\ 
\hline
TS3D~\cite{garbade2018two-ts3d} & RGB+D & (240, 60) & - & - & 76.1 & 25.9 & 93.8 & 48.9 & 33.4 & 31.2 & 66.1 & 56.4 & 31.6 & \textbf{38.5} & 51.4 & 30.8 & 46.2 \\ 
DDRNet~\cite{li2019rgbd-ddrnet} & RGB+D & (240, 60) & 88.7 & 88.5 & 79.4 & 54.1 & 91.5 & 56.4 & 14.9 & 37.0 & 55.7 & 51.0 & 28.8 & 9.2 & 44.1 & 27.8 & 42.8 \\
AIC-Net~\cite{li2020anisotropic} & RGB+D & (240, 60) & 88.2 & 90.3 & 80.5 & 53.0 & 91.2 & 57.2 & 20.2 & 44.6 & 58.4 & 56.2 & 36.2 & 9.7 & 47.1 & 30.4 & 45.8 \\ 
3D-Sketch~\cite{chen20203d} & RGB+D & (60, 60) & 90.6 & 92.2 & {84.2} & {59.7} & 94.3 & {64.3} & 32.6 & {51.7} & {72.0} & {68.7} & \textbf{45.9} & 19.0 & \textbf{60.5} & \textbf{38.5} &  {55.2} \\ 
IMENet~\cite{li2021imenet} & RGB+D & (60, 60) & 84.8 & 92.3 & {79.1} & {-} & - & {-} & - & {-} & {-} & {-} & {-} & - & {-} & - &  {47.5} \\ 
\hline
Ours & D & (60, 60) & \textbf{95.1} & 90.3 & \textbf{86.3} & \textbf{71.5} & {94.1} & \textbf{66.6} & {23.7} & \textbf{60.0} & \textbf{78.5} & \textbf{72.2} & {45.3} & {16.7} & {60.1} & {36.9} & \textbf{56.9} \\
\shline
\end{tabular}
}

\vspace{-0.35cm}
\end{center}
\caption{\textbf{Results on the NYUCAD dataset.} \textit{Resolution(a, b)} means the input resolution is $(a \times 0.6a \times a)$ and the output resolution is $(b \times 0.6b \times b)$.}
\label{tab:SotaOnNYUCAD}
\end{table*}

\begin{table*}[ht]
\begin{center}
\resizebox{\textwidth}{!}{
\begin{tabular}{l|c|c|c c c|c c c c c c c c c c c | c} 
\shline
& & & \multicolumn{3}{c|}{scene completion} & \multicolumn{12}{c}{semantic scene completion} \\ \hline
Methods & Input & Resolution & prec. & recall & IoU & ceil. & floor & wall & win. & chair & bed & sofa & table & tvs & furn. & objs. & mIoU \\ 
\hline
SSCNet~\cite{song2017semantic-sscnet} & D & (240,60) & 57.0 & \textbf{94.5} & 55.1 & 15.1 & 94.7 & 24.4 &  0.0 & 12.6 & 32.1 & 35.0 & 13.0 &  7.8 & 27.1 & 10.1 & 24.7\\
ESSCNet~\cite{zhang2018efficient-esscnet} & D & (240,60) & 71.9 & 71.9 & 56.2 & 17.5 & 75.4 & 25.8 &  6.7 & 15.3 & 53.8 & 42.4 & 11.2 &    0 & 33.4 & 11.8 & 26.7\\ 
VVNet~\cite{guo2018view-vvnet}  & D & (120,60) & 69.8 & 83.1 & 61.1 & 19.3 & 94.8 & 28.0 & 12.2 & 19.6 & 57.0 & 50.5 & 17.6 & 11.9 & 35.6 & 15.3 & 32.9  \\
ForkNet~\cite{wang2019forknet}  & D & (80,80) & - & - & 63.4 & 36.2 & 93.8 & 29.2 & 18.9 & 17.7 & {61.6} & 52.9 & 23.3 & 19.5 & {45.4} & 20.0 & 37.1 \\
CCPNet~\cite{zhang2019cascaded-ccpnet} & D & (240,240) & 74.2  & 90.8 & 63.5 & 23.5 & \textbf{96.3} & 35.7 & 20.2 & 25.8 & 61.4 & {56.1} & 18.1 & \textbf{28.1} & 37.8 & 20.1 & 38.5\\ 
SPCNet~\cite{zhong2020spcnet} & D & (240,60) & 72.1 & 42.2 & 36.3 & 33.8 & 64.4 & 38.3 & 7.5 & 30.7 & 53.4 & 42.6 & 19.7 & 5.5 & 34.2 & 13.9 & 31.3 \\ 
\hline
TS3D~\cite{garbade2018two-ts3d} & RGB+D & (240,60) & - & - & 60.0 & 9.7 & 93.4 & 25.5 & 21.0 & 17.4 & 55.9 & 49.2 & 17.0 & 27.5 & 39.4 & 19.3 & 34.1\\
SATNet~\cite{liu2018see-satnet} & RGB+D & (60,60) & 67.3 & 85.8 & 60.6 & 17.3 & 92.1 & 28.0 & 16.6 & 19.3 & 57.5 & 53.8 & 17.2 & 18.5 & 38.4 & 18.9 & 34.4 \\
DDRNet~\cite{li2019rgbd-ddrnet} & RGB+D & (60,60) & 71.5  & 80.8 & 61.0 & 21.1 & 92.2 & 33.5 & 6.8 & 14.8 & 48.3 & 42.3 & 13.2 & 13.9 & 35.3 &  13.2 & 30.4\\ 
AIC-Net~\cite{li2020anisotropic} & RGB+D & (60,60) & 62.4 & 91.8 & 59.2 & 23.2 & 90.8 & 32.3 & 14.8 & 18.2 & 51.1 & 44.8 & 15.2 & 22.4 & 38.3 & 15.7 & 33.3 \\ 
3D-Sketch~\cite{chen20203d} & RGB+D & (60,60) & {85.0} & 81.6 & {71.3} & {43.1} & 93.6 & {40.5} & {24.3} & {30.0} & 57.1 & 49.3 & {29.2} & 14.3 & 42.5 & {28.6} & {41.1} \\ 
IMENet~\cite{li2021imenet} & RGB+D & (60,60) & {90.0} & 78.4 & {72.1} & {43.6} & 93.6 & {42.9} & \textbf{31.3} & {36.6} & 57.6 & 48.4 & {32.1} & 16.0 & 47.8 & \textbf{36.7} & {44.2} \\ 
\hline
Ours & D & (60,60) & \textbf{91.1} & 79.7 & \textbf{74.0} & \textbf{51.4} & {94.0} & \textbf{49.9} & {15.9} & \textbf{41.9} & \textbf{68.3} & \textbf{58.8} & \textbf{35.4} & {12.9} & \textbf{48.5} & {29.1} & \textbf{46.0} \\
\shline
\end{tabular}
}
\vspace{-0.35cm}
\end{center}
\caption{\textbf{Results on the NYU dataset.} \textit{Resolution(a, b)} means the input resolution is $(a \times 0.6a \times a)$ and the output resolution is $(b \times 0.6b \times b)$.}
\label{tab:SotaOnNYU}
\end{table*}

\subsection{Implementation Details}
\label{sec:implementation_details}
We use the PyTorch framework with two Nvidia Titan Xp GPUs to conduct our experiments.
Mini-batch SGD with momentum of $0.9$ is adopted to train our network.
The initial learning rate is $0.05$, batch size is $8$ and the weight decay is $0.0005$.
We employ a Poly learning rate decay policy where the initial learning rate is multiplied by $(1-\frac{\text{now\_iter}}{\text{max\_iter}})^{0.9}$.
We train our network for $1000$ epochs on the NYUCAD dataset and the NYU dataset. 
The radius of the ellipsoidal receptive field is set to $0.09$ for the major axis and $0.03$ for the minor axes. 
We sample at most $8$ voxels inside each ellipsoidal receptive field. The balancing factor $\lambda$ in Equation~\ref{loss-func} is set to $0.5$.
Since the output of the SSC task is usually at the resolution of 60 $\times$ 36 $\times$ 60, we adopt the same input resolution of voxels following SATNet~\cite{liu2018see-satnet} and 3D-Sketch~\cite{chen20203d}.
Different from~\cite{zhong2020spcnet}, we feed both the observed and occluded points into our network during training. 
Since the number of points for each scene is not fixed, we randomly sample a fixed number of observed points ($2048$) and occluded points ($8192$) for each scene to enable batched training. 
At inference time, we use all the generated points as input.


\subsection{Comparisons with State-of-the-art Methods}

We compare the proposed method with state-of-the-art methods. 
Table~\ref{tab:SotaOnNYUCAD} lists the results on the NYUCAD dataset.
Our method outperforms all the existing voxel-based or point-based methods, gaining an increase of $1.7\%$ SSC mIoU and $2.1\%$ SC IoU compared to the previous best method~\cite{chen20203d}. 
While some methods use a higher input resolution, we already achieve good enough results with the 60 $\times$ 36 $\times$ 60 input resolution. 
The advantage of our method is not from a higher input resolution than others, but the novel and efficient point-voxel framework that modelling the local details and global context in a computationally-friendly manner.
We also conduct experiments on the NYU dataset to validate the performance of our method on realistic data.
As listed in Table~\ref{tab:SotaOnNYU}, our method consistently outperforms previous best method~\cite{li2021imenet} in both SC IoU and SSC mIoU metrics.
Notably, our method only requires one-pass forward, while IMENet~\cite{li2021imenet} performs multiple iterations between a 2D and a 3D network and introduces a very large computational cost.

\begin{figure}[ht]
    \centering
    \includegraphics[width=\linewidth]{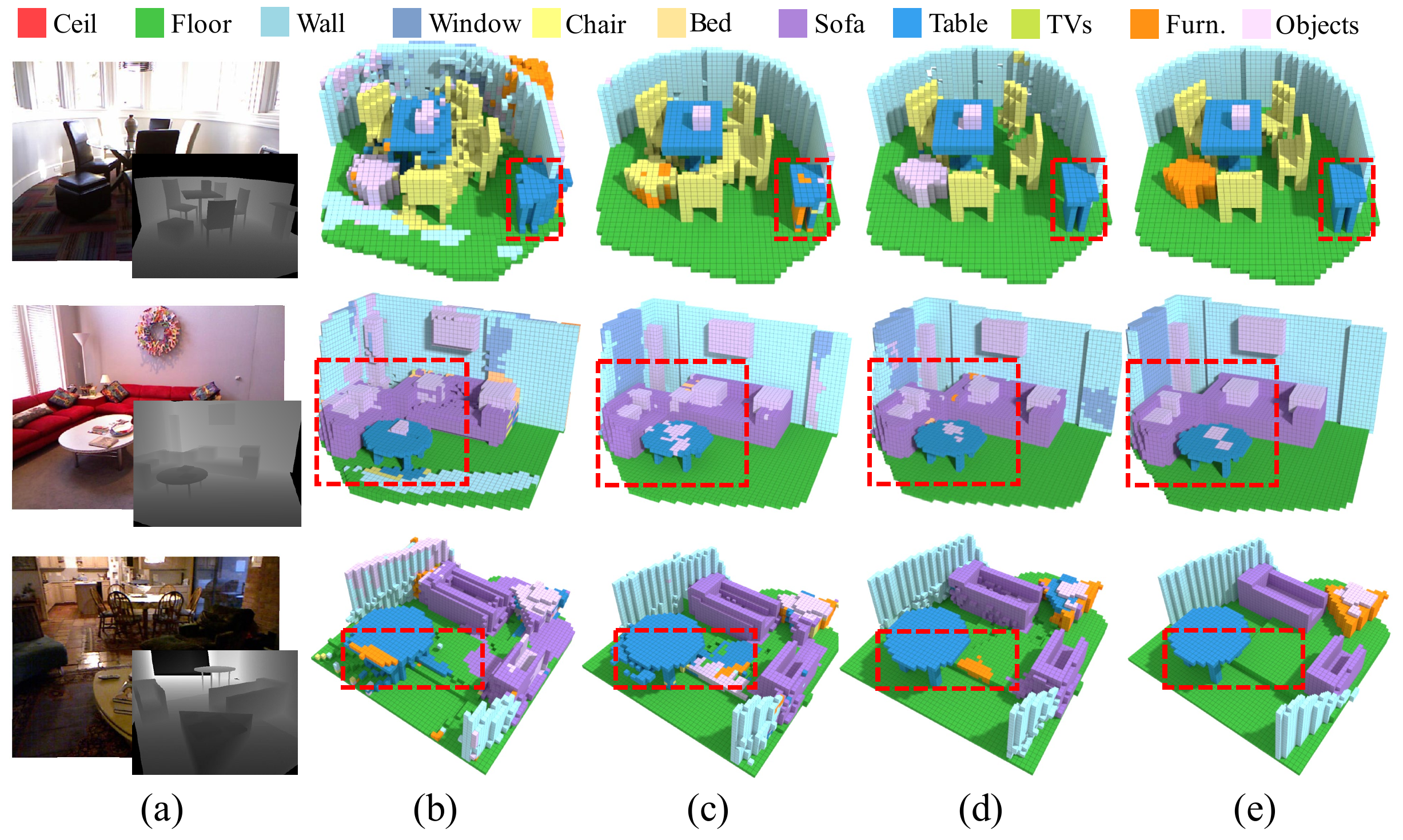}
    \caption{\textbf{Visualizations on the NYUCAD dataset.} From left to right: (a) RGB-D images, (b) results of~\protect\cite{song2017semantic-sscnet}, (c) results of~\protect\cite{chen20203d}, (d) our results, and (e) ground truth. Best viewed in color and zoom in.}
    \label{fig:qualitive_nyucad}
\end{figure}

We provide some visualizations on the NYUCAD dataset in Figure~\ref{fig:qualitive_nyucad}.
With only the depth images as input, our method achieves good inter-class distinction and intra-class consistency.
We think the superiority of the proposed method comes from the two-stream framework, where the point stream extracts the high-level semantics and the voxel stream extracts the detailed local structure information.

\begin{table}[ht]
    \small
    \centering
    \setlength{\tabcolsep}{5pt}
    \begin{tabular}{lcccc}
        \shline
        Methods & FLOPs & Memory & SC IoU(\%) & SSC mIoU(\%) \\
        \hline
        SSCNet & 163.8G & 1057M & 73.2 & 40.0 \\
        3D-Sketch & 293.7G & 1535M & 84.2 & 55.2 \\
        Ours   & 8.9G   & 554M  & 86.3 & 56.9 \\
        \shline
    \end{tabular}
    \caption{{Efficiency analysis on the NYUCAD dataset.}}
    \label{tab:efficiency_analysis}
\end{table}

\subsection{Efficiency Analysis} 
Since the computational cost of our method depends on the number of visible empty points in each scene, we report the average number of these statistics on the NYUCAD test set in Table~\ref{tab:efficiency_analysis}. 
The proposed method avoids the redundancy caused by the visible empty voxels and achieves higher performance with lower computational cost. 
Even though our FLOPs are only $5.4\%$ of SSCNet and $3.0\%$ of 3D-Sketch, we still achieve a better performance.


\subsection{Ablation Study}
In this section, we conduct ablation studies to verify the effectiveness of each aforementioned component.
All the results are tested on the NYUCAD test set.

\paragraph{Proposed modules.} 
Firstly, we do ablation studies on different modules in our method in Table~\ref{tab:ablation_modules}. 
The AVA module aggregates the local structure information from the voxel stream to the point stream, and the SP module encourages feature propagation of points belonging to the same category. 
Adopting either of these two modules improves the performance, and the combination of them achieves the maximum benefits.
Note that our point-based baseline already outperforms some of the voxel-based methods. 
Thanks to the removal of redundant parts in the input, the point stream could have a deeper architecture and higher feature dimensions, leading to stronger representation power.


\begin{table}[ht]
    \small
    \centering\setlength{\tabcolsep}{9pt}
    
    \begin{tabular}{ccccc}
        \shline
        Baseline   & AVA        & SP             & SC IoU(\%)   & SSC mIoU(\%) \\
        \hline
        \checkmark &            &                & 85.1           & 51.4            \\
        \checkmark &            &    \checkmark  & 85.9           & 53.2 \\
        \checkmark & \checkmark &                & 85.3           & 54.9 \\
        \checkmark & \checkmark &    \checkmark  & \textbf{86.3}  & \textbf{56.9} \\
        \shline
    \end{tabular}
    
    \caption{\textbf{Ablation study on different modules.} The baseline is a pure point-based network (PointNet++\protect\cite{qi2017pointnet++}), `AVA' means Anisotropic Voxel Aggregation and `SP' means Semantic-aware Propagation.
        }
        \label{tab:ablation_modules}
        \vspace{-3mm}
\end{table}
\begin{table}[ht]
    \small
    \centering\setlength{\tabcolsep}{6pt}
    
    \begin{tabular}{lcc}
        \shline
        Methods & SC IoU(\%) & SSC mIoU(\%) \\
        \hline
        Point Only & 85.9 & 53.2  \\ 
        Nearest Aggregation & 86.0 & 55.1  \\
        Spherical Aggregation & 86.1 & 56.0  \\
        Anisotropic Voxel Aggregation & \textbf{86.3} & \textbf{56.9} \\
        \shline
    \end{tabular}
    \caption{{Ablation study on different Voxel Aggregation Strategies.}}
    \label{tab:ablation_pvi}
    \vspace{-4mm}
\end{table}

\paragraph{Voxel Aggregation Strategies.} 
As illustrated before, we propose the AVA module that aggregates the voxel structure features in an anisotropic manner. 
Here, we try to compare AVA with other voxel aggregation strategies to verify its effectiveness.
We provide two other strategies: `Nearest Aggregation' and `Spherical Aggregation'. 
`Nearest Aggregation' means we only concatenate the point features with the features of the nearest voxel from the point. 
`Spherical Aggregation' means we adopt a spherical receptive field with the radius $r$ to aggregate the voxel features inside the receptive field. 
$r$ is $0.09$ here so that it is the same with the radius of the major axis in AVA. 
Results are listed in Table~\ref{tab:ablation_pvi}. 
As shown in the table, both `Nearest Aggregation' and `Spherical Aggregation' could boost the performance, because they more or less introduce some detailed geometric information to the point stream. However, the proposed AVA module achieves the best performance with the anisotropic aggregation design, because it could capture more feature patterns in different directions.

\paragraph{Position of the  AVA Module.} 
We try to embed AVA modules to different positions in the network, as listed in Table~\ref{tab:ablation_pos}. 
We find that if we embed the AVA module to a higher level layer in the network, the additional gain brought by voxel features decreases. 
This makes sense for that the AVA module exploits local features in the voxel representation and if we embed it into the first SA layer, 
the structure information provided by the AVA module will be further encoded as the network goes deeper. 
Also, if we embed AVA modules to all the SA layers in the network, the performance is just similar to the proposed method.
Hence, we only embed it to the first SA layer for lower computational cost.

\paragraph{Feature Propagation Strategies.} 
We then conduct experiments on the proposed SP module. 
The SP module encourages feature propagation in points belonging to the same category through the $w_{i,j}$ defined in Equation~\ref{eq:edge-weight} and Equation~\ref{eq:pair-sim}. 
We compare SP with other feature propagation strategies in Table~\ref{tab:ablation_fp}.
Our SP achieves the best performance, since it considers pairwise semantic relations and avoids the errors caused by feature propagation from the wrong categories.

\begin{table}[ht]
    \small
    \centering\setlength{\tabcolsep}{6pt}
    
    \begin{tabular}{ccccccc}
        \shline
        SA-1 & SA-2 & SA-3 & SA-4 & SC IoU(\%) & SSC mIoU(\%) \\
        \hline
        \checkmark & & & & \textbf{86.3} & \textbf{56.9}  \\
        & \checkmark & & & 86.2 & 55.7 \\
        & & \checkmark & & 86.0 & 53.4 \\
        & & & \checkmark & 85.8 & 53.8 \\
        \checkmark & \checkmark & \checkmark & \checkmark & 86.2 & 56.6 \\
        \shline
    \end{tabular}
    \caption{\textbf{Ablation study on the position of AVA module.} SA-$i$ indicates AVA module is embedded into the $i$-th SA layer.}
    \label{tab:ablation_pos}
    \vspace{-4mm}
\end{table}
\begin{table}[ht]
    \small
    \centering
    
    \begin{tabular}{lcc}
        \shline
        Methods & SC IoU(\%) & SSC mIoU(\%) \\
        \hline
        Inverse Euclidean & 85.3 & 54.9  \\
        Cosine Similarity & 83.9 & 54.4  \\
        Semantic-aware Propagation & \textbf{86.3} & \textbf{56.9}  \\
        \shline
    \end{tabular}
    \caption{\textbf{Ablation study on different Feature Propagation Strategies.}
        `Inverse Euclidean' means the inverse Euclidean distance between two points used in Pointnet++~\protect\cite{qi2017pointnet++}. 
        `Cosine Similarity' means the cosine similarity between features of the two points.
        The proposed SP module achieves the best performance.
    }
    \label{tab:ablation_fp}
    \vspace{-3mm}
\end{table}

\section{Conclusion}
\label{sec:conclusion}
In this paper, we introduce the Point-Voxel Aggregation Network for Semantic Scene Completion, 
which combines the advantages of the computationally efficient point representation and the rich-detailed voxel representation. Besides, an AVA module is proposed to aggregate the structure information from voxels and a SP module is proposed to encourage semantic-aware feature propagation in points.
Experimental results demonstrate the effectiveness and efficiency of our method with state-of-the-art performance on two public benchmarks, with only the depth images as input. 
In the future, we believe the representation of the scene is crucial for this topic, such as the modality of inputs (\eg, RGB and depth map), as well as the point cloud with different sampling rates.

\section*{Acknowledgements}
This work is supported by the National Key Research and Development Program of China (2017YFB1002601), National Natural Science Foundation of China (61632003, 61375022, 61403005), Beijing Advanced Innovation Center for Intelligent Robots and Systems (2018IRS11), and PEK-SenseTime Joint Laboratory of Machine Vision.

\bibliography{ref}

\end{document}